# Effective Features of Remote Sensing Image Classification Using Interactive Adaptive Thresholding Method

T. Balaji [1], M. Sumathi [2]

[1] Assistant Professor, Dept. of Computer Science, Govt. Arts College, Melur, India
[2] Associate Professor, Dept. of Computer Science, Sri Meenakshi Govt. Arts College for Women, Madurai, India
[1] bkmd_gacm@rediffmail.com
[2] sumathivasagam@gmail.com

*Abstract* - Remote sensing image classification can be performed in many different ways to extract meaningful features. One common approach is to perform edge detection. A second approach is to try and detect whole shapes, given the fact that these shapes usually tend to have distinctive properties such as object foreground or background. To get optimal results, these two approaches can be combined. This paper adopts a combinatorial optimization method to adaptively select threshold based features to improve remote sensing image. Feature selection is an important combinatorial optimization problem in the remote sensing image classification. The feature selection method has to achieve three characteristics: first the performance issues by facilitating data collection and reducing storage space and classification time, second to perform semantics analysis helping to understand the problem, and third to improve prediction accuracy by avoiding the "curse of dimensionality". The goal of this thresholding an image is to classify pixels as either dark or light and evaluation of classification results. Interactive adaptive thresholding is a form of thresholding that takes into account spatial variations in illumination of remote sensing image. We present a technique for remote sensing based adaptive thresholding using the interactive satellite image of the input. However, our solution is more robust to illumination changes in the remote sensing image. Additionally, our method is simple and easy to implement but it is effective algorithm to classify the image pixels. This technique is suitable for preprocessing the remote sensing image classification, making it a valuable tool for interactive remote based applications such as augmented reality of the classification procedure.

*Keywords* – Simple Thresholding, Colour Thresholding, Region, Clustering, Matching, Edge Based Classification, AT(Adaptive Threshold) and BG(Background).

I. INTRODUCTION

Image thresholding is a common task in many computer vision, graphics and remote sensing applications. Image thresholding classification of remote sensing image based on a certain characteristic of the pixels (for example, intensity value). The goal is to create a binary representation of the image and classifying each pixel into one of two categories, such as dark or light. This is a common task in many image processing applications and some computer graphics applications. The most basic thresholding method is to choose a fixed threshold value and compare each pixel to that value. These techniques have been described and evaluated extensively in a number of threshold values. However, fixed thresholding often fails if the illumination varies spatially in the image or over time in a image scene.

In order to account for variations in illumination, the common solution is to take adaptive thresholding method. The main difference is that a different threshold value is computed for each pixel in the image. This technique provides more robustness to changes in illumination features. A number of adaptive thresholding methods exist algorithms that use more than one threshold values, which enables to assign pixels to one of a few classes. Threshold values may be entered manually or automatically.

The proposed classification approach is conceptually different and explores a new strategy. In fact, instead of considering only one remote sensing image for each application, our technique consists in combining many realizations of the same image, in order to increase the information quality and to get an optimal classified image. This work is a very simple and clear adaptive technique using remote sensing images. This method is easy to implement for interactive performance on a remote sensing image stream. It often provides sufficient accuracy and high processing speed to increase robustness to strong illumination changes. In addition, we present a clear and tidy solution without increasing the complexity of the implementation. However this algorithm is found to be reliable and robust to different kind of remote sensing images. The adaptive method provides an alternative way of selecting the dominant peaks of the image histogram according to predefined constraints and it makes no assumptions and suitable for use in any application. This proposed classification approach is conceptually different and explores a new strategy. In fact, instead of considering only one remote sensing image for each application, our technique consists in combining many realizations of the same image, in order to increase the information quality and to get an optimal classified image. The classification accuracy of the proposed method is evaluated and a comparative study versus existing techniques is presented an extensive set of remote sensing colour images. Satisfactory for classification results have been obtained showing the effectiveness and superiority of the adaptive method.

II. CLASSIFICATION OVERVIEW

Classification is a partitions an image into distinct regions containing each pixel with similar attributes. It is to be meaningful and useful for image analysis and interpretation should strongly relate to depicted objects or features of interest. Meaningful classification is the first step from low-level image processing transforming a grayscale or color image into one or more other remote sensing images to high-level image description in terms of features, objects, and scenes. The success of image analysis depends on reliability of classification, but an accurate partitioning of an image is generally a very challenging problem of remote sensing image applications. Classification techniques are either contextual or non-contextual. There is no account of spatial relationships between features in an image and group pixels together on the basis of some global attribute, e.g. gray level or color. Contextual techniques additionally exploit these relationships, e.g. group together pixels with similar gray levels and close spatial locations.The division of remote sensing image into meaningful structures and image classification is often an essential step in image analysis, object representation, visualization and many other image processing tasks. A great variety of classification methods has been proposed in the past decades and some categorization is necessary to present the methods properly derived. The categorization presented in this paper is therefore rather a categorization regarding the emphasis of an approach than a strict division. The following categories are used to classifying the remote sensing image:

A. Threshold Based Classification

Histogram thresholding and slicing techniques are used to classifying the image. This may be applied directly to an image, but can also be combined with pre and post-processing techniques.

B. Edge Based Classification

With this technique, detected edges in an image are assumed to represent object boundaries and used to identify these objects.

C. Region Based Classification

Where an edge based technique may attempt to find the object boundaries and then locate the object itself by filling them inner boundaries, a region based technique takes the opposite approach, by starting in the middle of an object and then growing outward until it meets the object boundaries.

D. Clustering Based Classification

In this context, clustering methods attempt to group together patterns that are similar in some sense of remote sensing image. This goal is very similar to what we are attempting to do when we classify an image, and indeed some clustering techniques can really be applied for remote sensing image classification.

E. Matching Based Classification

When we know what an object we wish to identify in an image (approximately) looks like, we can use this knowledge to locate the object in an image. This approach to classification of an image is called matching.

The perfect image classification in each pixel is assigned to the correct object classification. This is a goal that cannot usually be achieved. Indeed, because of the way a remote sensing image is acquired, this may be impossible. Since a pixel may straddle the real boundary of objects such that it partially belongs to two (or even more) objects. Most current classification methods, attempt to assign a pixel to a single region, which is an approach that is more than adequate for some applications. Methods that assign a classifying probability distribution to each pixel is called probabilistic. This class of methods is theoretically more accurate, and applications where a probabilistic approach is the only approach accurate enough for specific object measurements can easily be named. Perfect image classification is also often not reached because of the occurrence of over classification or under classification. In the first case, pixels belonging to the same object are classified as belonging to different classification. A single object may be represented by two or more classification. In this case pixels belonging to different objects are classified as belonging to the same object. A single classification may contain several objects foreground and background features.

The goal is to create classification algorithm with the following characteristics:

- It should be able to separate foreground objects in varying brightness conditions.
- Separated foreground objects should have smooth and accurate edges.
- Noise immunity would be an advantage.
- Possible computationally fast implementation.

III. THRESHOLDING TECHNIQUES

Thresholding is the effective and simplest non-contextual classification technique for gray scale or color based image. With a single threshold, it transforms a grayscale or color image into a binary image considered as a binary region map. The binary map contains two possibly disjoint regions, one of them containing pixels with input data values smaller than a threshold and another relating to the input values that are at or above the threshold. The former and latter image regions are usually labeled with zero (0) and non-zero (1) labels respectively. The classification depends on image property being thresholded and on how the threshold is chosen.

Generally, the non-contextual thresholding may involve two or more thresholds as well as produce more than two types of regions such that ranges of input image signals related to each region type are separated with thresholds. Some thresholding technique is automatically determining the threshold value.

**A. Simple Thresholding**

The most common image property to threshold is pixel gray level is calculated as:

$g(x, y) = 0$  if $f(x, y) < T$ and

$g(x, y) = 1$  if $f(x, y) \geq T$

Where T is the threshold value of the given image. There are two thresholds using, T1 and T2, a range of gray levels related to region 1 can be defined as two cases:

Case 1: If $f(x, y) < T1$ or $f(x, y) > T2$ then $g(x, y) = 0$

Case 2: If $T1 \leq f(x, y) \leq T2$ then $g(x, y) = 1$

A general approach to thresholding is based on assumption that images are multimodal, that is, different objects of interest relate to distinct peaks (or modes) of the 1-D signal histogram or 2-D signal histogram. The thresholds have to optimally separate these peaks in spite of typical overlaps between the signal ranges corresponding to individual peaks. The optimal threshold that minimizes the expected numbers of false detections and rejections may not coincide with the lowest point in the valley between two overlapping peaks.

**B. Colour Thresholding**

Colour classification may be more accurate because of more information at the pixel level comparing to gray scale images. The standard Red-Green-Blue (RGB) colour representation has strongly interrelated colour components, and a number of other colour systems (e.g. HSI Hue-Saturation-Intensity) have been designed in order to exclude redundancy, determine actual object / background colours irrespectively of illumination, and obtain more stable classification. Classification of colour images involves a partitioning of the colour space, i.e. RGB or HSI space. One simple approach is based on some reference (or dominant) colour ($R_0$, $G_0$, $B_0$) and thresholding of Cartesian distances to it from every pixel colour is calculated as:

$$f(x, y) = (R(x, y), G(x, y), B(x, y))$$

Where $g(x,y)$ is the binary region map after thresholding. This thresholding rule defines a sphere in RGB space, centered on the reference colour. All pixels inside or on the sphere belong to the region indexed with 1 and all other pixels are in the region 0. A colour histogram is built by partitioning of the colour space onto a fixed number of bins such that the colours within each bin are considered as the same colour. Generally, a good complete classification of image must satisfy the following criteria:

- All pixels have to be assigned to regions.
- Each pixel has to belong to a single region only.
- Each region is a connected set of pixels.
- Each region has to be uniform with respect to a given predicate.
- Any merged pair of adjacent regions has to be non-uniform.

## IV. PROPOSED INTERACTIVE ADAPTIVE THRESHOLDING METHOD

The threshold method separates the background from the object, the adaptive separation may take account of empirical probability distributions of object (e.g. dark) and background (bright) pixels. Such a threshold has to equalize two kinds of expected errors are assigning a background pixel to the object and of assigning an object pixel to the background. A simple interactive adaptation threshold is based on successive refinement of the estimated peak positions. It assumes that two elementary aspects:

- Each peak coincides with the mean gray level for all pixels that relate to that peak
- The pixel probability decreases monotonically on the absolute difference between the pixel and peak values both for an object and background peak.

Remote sensing image classification is a tool that can be used whenever we have a function from pixels to real numbers $f(x,y)$ (pixel intensity) and we wish to compute the sum of this function over a rectangular region of the image. However, if we need to compute the sum over multiple overlapping rectangular windows, we can use remote sensing image and achieve a constant number of operations per rectangle with only a linear amount of preprocessing. To compute the image, we store at each location is $I(x,y)$, the sum of all $f(x,y)$ terms to the left and above the pixel $(x,y)$. This is accomplished in linear time using the following equation for each pixel (taking into account the border cases):

$$I(x, y) = f(x, y) + I(x-1, y) + I(x, y-1) - I(x-1, y-1)$$

The sum of the function for any rectangle with upper left corner $(x_1, y_1)$, and lower right corner $(x_2, y_2)$ can be computed in constant time using the following equation:

$$\sum_{x=x_1}^{x_2} \sum_{y=y_1}^{y_2} f(x,y) = I(x_2, y_2) - I(x_2, y_1-1) - I(x_1-1, y_2) + I(x_1-1, y_1-1)$$

The proposed adaptive thresholding technique is a simple extension of global threshold method. The main idea of this algorithm is that each pixel is compared to an average of the surrounding pixels. Specifically, an approximate moving average of the last S pixels seen is calculated while traversing the image. If the value of the current pixel is T percent lower than the average then it is set to black, otherwise it is set to white. This method works because comparing a pixel to the average of nearby pixels will preserve hard contrast lines and ignores soft gradient changes.

A. Methodology

The proposed adaptive threshold method is able to operate in varying brightness conditions and our custom threshold function will be referred as adaptive threshold in remote sensing image. This algorithm consists of the following steps:

1. Input remote sensing image $R_{IN}$.
2. Calculate global threshold value $T_G$.
3. Take the pixel $P_i$.
4. Calculate brightness value $I_i$ in pixel window $W_{PI}$.
5. Calculate adaptive local threshold value $T_L$ in threshold window $W_T$.
6. Calculate absolute difference $A_D$ between $I_i$ and $T_L$.
7. Assign pixel binary value $B_i$:
   Use adaptive global threshold $T_G$ if $A_D$ is less than sensitivity S, else use local adaptive threshold $T_L$. Set $B_i$ as foreground if $I_i$ is greater or equal to threshold value.
8. Repeat steps 3 to 7 until every pixel $P_i$ has been assigned a binary value $B_i$.
9. Output binary image $R_{OUT}$.

The advantage of this method is that only a single pass through the image is required. However, this method is that it is dependent on the scanning order of the pixels. In addition, the moving average is not a good representation of the surrounding pixels at each step because the neighborhood samples are not evenly distributed in all directions. Our technique is clean, straightforward, easy to code, and produces the same output independently of how the image is processed. Instead of computing a running average of the last S pixels seen, we compute the average of an S x S window of pixels centered around each pixel. This is a better average for comparison since it considers neighboring pixels on all sides. The average computation is accomplished in linear time by using the remote sensing image. We calculate the interactive image in the first pass through the input image. In a second pass, we compute the S x S average using the remote sensing image for each pixel in constant time and then perform the comparison. If the value of the current pixel is T percent less than this average then it is set to black, otherwise it is set to white. The following pseudo code demonstrates our adaptive technique for input image **in**, output binary image **out**, image width **w** and image height **h**.

Adaptive threshold method based on multiple average value calculations and their differences. The proposed algorithm has three tunable values: $W_P$ is the length of pixel window side, $W_T$ is the length of threshold window side, S is sensitivity value. There are two windows of adjustable sizes. Average value in smaller window (pixel window $W_P$) defines its centre pixel brightness value I. It is expressed as arithmetic mean of all pixel intensities in the window:

$$I = 1/n \sum_{i=0}^{n} P_i$$

Where $n = W_P^2$ is the pixel count in small window and $P_i$ is $i^{th}$ pixels intensity value in window $W_P$. The actual procedure of the proposed method is shown below:

This algorithm works initially parsing every pixel in the image and storing integral sum (running sum) which depends on upper and left neighbor pixels. Average value in larger window (threshold window $W_T$) defines local threshold value $T_L$, which is also calculated as arithmetic mean of pixel intensities in larger window is given by the equation:

$$T_L = 1/m \sum_{j=0}^{m} P_j$$

where $m = W_T^2$ denotes pixel count in threshold window $W_T$. The sensitivity S represents lowest reliable value of absolute difference between pixel intensity value I and local threshold value $T_L$. If value is higher or equal than S then we have a reliable calculation of foreground or background pixel. If absolute difference is lower than S then threshold value of such pixel is calculated using expression in third line of the following equation:

$$B = \begin{cases} 0, \text{ when } I < T_L \text{ and } |I - T_L| \geq S \\ 1, \text{ when } I \geq T_L \text{ and } |I - T_L| \geq S \\ B_{VA}, \text{ when } |I - T_L| < S \end{cases}$$

Where $B$ is final pixel binary value after threshold operation, $B_{VA}$ indicates binary value assignment by comparison to arithmetic mean of all pixel intensity values of complete image $T_G$ is given by the equation:

$$B_{VA} = \begin{cases} 0, \text{ when } I < T_G \\ 1, \text{ when } I = T_G \end{cases}$$

B. Procedure

We choose to create a new specific adaptive threshold function, suitable for our needs in remote sensing image classification. Interactive adaptive thresholding involves selecting a good gray scale at which to re-quantize an image to a small number of gray scales. We will only discuss interactive thresholding an algorithm in remote sensing image i.e. the objective is to create an image of just black and white. Adaptive thresholding algorithms operate either globally or locally. This algorithm has the following comprehensive procedure:

```
Procedure AdaptiveThreshold (Rin, Rout, w, h)
For i = 0 to w do
    Sum ← 0
    For j = 0 to h do
        Sum ← Sum + Rin[i,j]
        If i = 0 then
            Rsimg[i, j] ← Sum
        Else
            RsImg[i, j] ← Rsimg[i−1, j] + Sum
        End if
    End for
End for
For i = 0 to w do
    For j = 0 to h do
        x1 ← i − S/2
        x2 ← i + S/2
        y1 ← j − S/2
        y2 ← j + S/2
        count ← (x2−x1) × (y2−y1)
        sum ← { Rsimg[x2,y2] − Rsimg[x2,y1−1] −
                Rsimg[x1−1,y2] + Rsimg [x1−1,y1−1]
        If (in[i, j] × count) ≤ (sum × (100−t) / 100) then
            Rout[i,j] ← 0
        Else
            Rout[i,j] ← 255
        End if
    End for
End for
```

C. Innovations

Our algorithm introduces several innovations to classic or interactive remote sensing images:

- It uses local average intensity value not only for threshold calculation but also for pixel value calculation, which provides smoother edge lines to classified remote sensing objects.
- Sensitivity parameter S is the solution for continuous intensity regions larger that local threshold window, which otherwise would introduce unnecessary objects into classified image.
- It introduces speed optimization based on interactive remote sensing image which enables fast adaptive classification applications.

## V. EXPERIMENTAL RESULTS

This section presents the results obtained from IATM using robust and powerful thresholding algorithm which was carried out for the experiments. This type of remote sensing image classification was implemented using MATLAB version 7.6 software. The complete process of this algorithm standards are summarized in the following fig. 2.

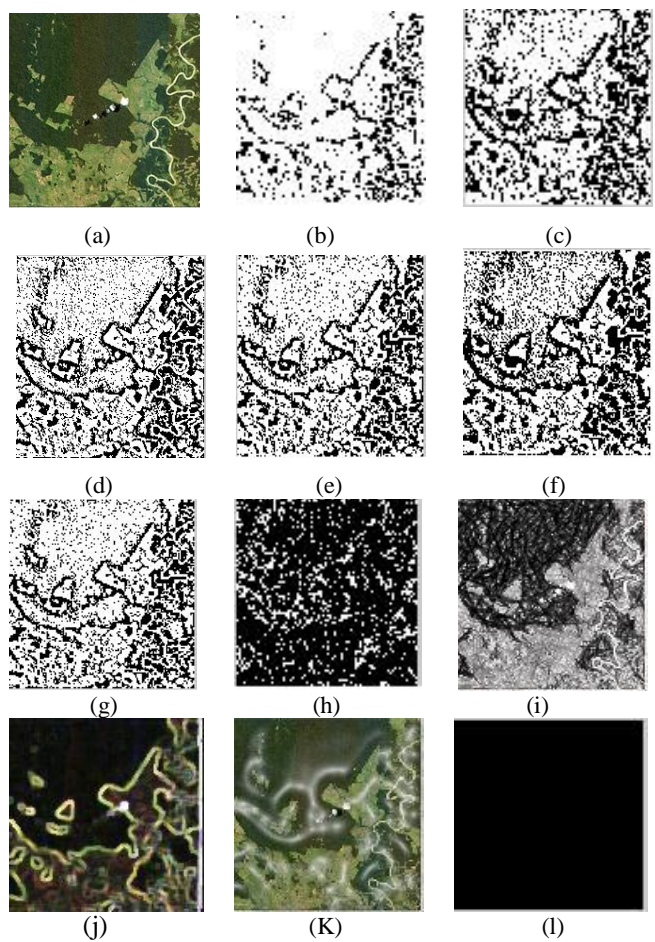

Figure 2. a. Original Satellite Image, b. Gray Scale AT First Frame, c. Gray Scale AT Second Frame, d. Gray Scale AT Third Frame, e. Gray Scale AT Fourth Frame, f. Gray Scale AT Fifth Frame, g. Gray Scale AT Sixth Frame, h. Subtracted BG Threshold, i. Artistic BG Threshold, j. Adaptive BG Threshold Glow Edges, k. Adaptive BG Threshold Wrapping Edges, l. Subtracted BG Binary Image.

## VI. CONCLUSION

This algorithm is suitable for remote sensing image classification application because its precision is high enough and error rate in classified images is very low. High precision in combination with low calculation complexity due to sum satellite application allowed us to perform classification of thousands of images with very successful results. Adaptive classification algorithm was proposed, implemented and experimentally tested acceptable times in the usual applications. Our proposed adaptive classification algorithm gives standard deviation five times smaller than ordinary bi-level classification algorithm. This algorithm was optimized for computation speed 10.58912, which is the combination of high classification precision, makes it suitable for remote sensing applications. This algorithm is clearly shows that give object edges much more similar and close to ideal classified object edges. All the statistical parameters show that objects obtained using our adaptive threshold classification algorithm are reliably better in comparison to bi-level thresholded objects data. To compare classification accuracy, difference between each ideal contour pixel and corresponding edge pixel in classified image was calculated. In our case corresponding classified edge pixels are the closest ones to the ideal contour pixels.